\documentclass{article} %
\usepackage[preprint]{colm2026_conference} %

\usepackage{microtype}
\usepackage{hyperref}
\usepackage{url}
\usepackage{booktabs}
\usepackage{tabularx}
\usepackage{multirow}
\usepackage{fancyvrb}
\usepackage{graphicx}
\usepackage{float}
\usepackage{xcolor}
\definecolor{ours}{HTML}{1F5FBF}
\definecolor{parallel}{HTML}{E0A100}
\definecolor{single}{HTML}{8A8F99}
\definecolor{pct}{HTML}{616E7C}
\definecolor{better}{HTML}{3B7A57}  %
\definecolor{worse}{HTML}{A8463F}   %
\usepackage{lineno}

\definecolor{darkblue}{rgb}{0, 0, 0.5}
\hypersetup{colorlinks=true, citecolor=darkblue, linkcolor=darkblue, urlcolor=darkblue}

\usepackage{amsmath,amsfonts,bm}

\def\eqref#1{equation~\ref{#1}}

\def\1{\bm{1}}

\DeclareMathAlphabet{\mathsfit}{\encodingdefault}{\sfdefault}{m}{sl}
\SetMathAlphabet{\mathsfit}{bold}{\encodingdefault}{\sfdefault}{bx}{n}

\title{Scaling Discovery Requires Forgetting}

\author{Shangyin Tan$^1$, Ben Zhou$^1$, Lakshya A Agrawal$^1$, Jialin Zhang$^1$, Yilin Chen$^2$, \\
\bf Qiuyang Mang$^1$, Donghyun Lee$^1$, Dennis Jacob$^1$, Koushik Sen$^1$, Alex Dimakis$^{1,3}$, \\ \bf Matei Zaharia$^1$ \\
$^1$UC Berkeley \quad $^2$Stanford \quad $^3$Bespoke Labs}

\definecolor{ailightnavy}{RGB}{140,170,215}

\begin{document}

\ifcolmsubmission
\linenumbers
\fi

\maketitle
\fancyhead{}\renewcommand{\headrulewidth}{0pt} %

\begin{figure}[H]
  \centering
  \includegraphics[width=\textwidth]{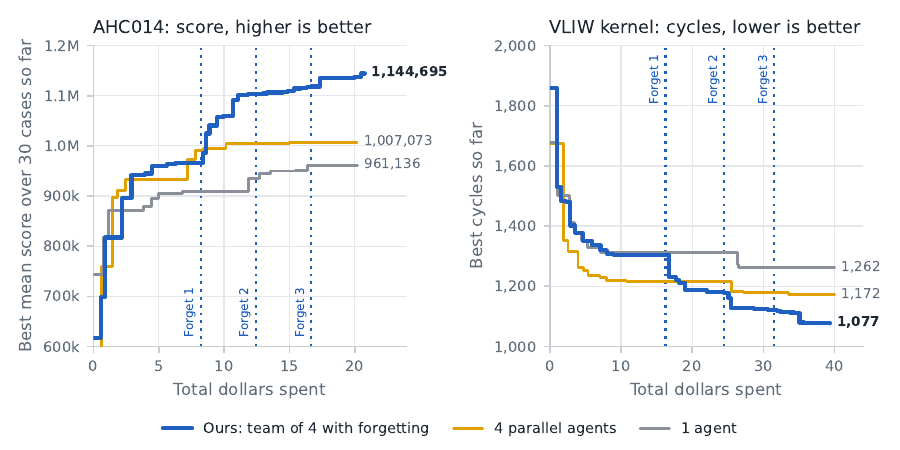}
  \caption{Best score so far against total money spent on AtCoder Heuristic Contest 014 and Anthropic's VLIW kernel take-home, for the best replicate of each configuration. \textcolor{single}{\textbf{One autoresearch agent}} with the same session continued throughout and \textcolor{parallel}{\textbf{four parallel agents}} both plateau early. \textcolor{ours}{\textbf{A team of four agents}} plateaus as well, until each dotted line, where every state and session is discarded, and the team restarts from the best candidate; each forget unblocks the plateau and the score climbs again. All three use the same configuration and have the same total money spent.}
  \label{fig:main}
\end{figure}

\begin{abstract}
    Automated research with LLM agents has shown promising results in various domains. Through iterative refinement and scored verifiers, LLM agents are capable of discovering new ideas.
    There are different ways to scale automated research: let one agent iterate for longer, run several agents in parallel, or further let the parallel agents communicate and build on one another's candidates.
    However, we find that all three strategies eventually plateau as the agents accumulate history. Each agent becomes anchored to its own past attempts, making fewer and smaller edits. Beyond this point, additional budget brings little further improvement.
    In this paper, we show that \textit{the key to scaling discovery further is to forget}. Periodically discarding the search state, including every agent's session and the candidate history, and restarting from the best candidate found so far, unblocks the search and pushes the same budget further.

    We implement forgetting in a team of agents with communication and evaluate it on four tasks spanning heuristic programming, kernel optimization and model training. The results show up to 15\% improvement compared to parallel agents, and up to 39\% compared to a single agent.
    Our results suggest that, for open-ended discovery, what an agent is allowed to remember matters as much as how much compute it is given.
\end{abstract}

\section[Introduction]{Introduction}\label{sec:intro}

Automated research and discovery increasingly rely on iterative refinement, where a system proposes a solution, measures it, learns from the outcome, and proposes a better one. Recently, large language model (LLM) agents have automated this cycle end-to-end. These autoresearch agents have produced new results in mathematics, algorithm design, prompt optimization, and machine learning research \citep{agrawal2026gepa,novikov2025alphaevolve,karpathy2026autoresearch}. As the underlying models grow more capable, autoresearch agents can now sustain this process for hours on a single problem \citep{liu2026agentsslowdownunderstanding}.

With more compute and time, an autoresearch agent can be scaled in three ways. One is to run a single agent longer, giving it more opportunities to refine its ideas. Another is to run multiple agents in parallel, exploring a broader set of ideas at once. A third is to allow those agents to share their progress, so that an idea discovered by one can be developed by the others. These three strategies mirror ways of scaling a human research effort.

However, in practice none of them scales indefinitely. We find that all three strategies eventually plateau, limited by the same underlying mechanism. In every strategy, each agent carries its growing autoresearch state and session context into every new proposal. As this history accumulates, the agent's proposals become increasingly conservative, and the agent's progress stalls well before the budget is exhausted (Figure~\ref{fig:main}).

We introduce a key concept to address this plateau: \emph{forgetting}. When an agent forgets, it discards its session, along with all candidate solutions and records in the filesystem state, and keeps only the best candidate. A new agent with an empty context then starts from this best candidate.
Forgetting has two benefits. First, little of the progress made is lost. We observe that in iterative autoresearch, the best candidate often subsumes many effective heuristics from earlier iterations. Second, the new agent is no longer conditioned on earlier failed attempts.

Forgetting works well with communication. In multi-agent team settings, agents build on each other's candidates, so a good idea found by one agent quickly propagates to the others. We periodically make the whole team forget, where every agent reinitializes from the single best candidate. This clears the accumulated history before it constrains fresh exploration (Figure~\ref{fig:idea}).

We evaluate \emph{forgetting} on four tasks: writing a heuristic solver for a programming contest problem \citep[AtCoder Heuristic Contest 014;][]{atcoder2022ahc014}, solving an algorithmic optimization problem \citep[Frontier-CS;][]{mang2025frontiercsevolvingchallengesevolving}, optimizing a VLIW kernel \citep[Anthropic Kernel Challenge;][]{anthropic2026performancetakehome}, and improving a PyTorch script for training a small language model \citep[nanochat-autoresearch;][]{karpathy2026autoresearch}. A team of four agents with forgetting is best on every task and improves on the strongest baseline by 1.9\% to 15\%. These results show that \textbf{forgetting unblocks plateaued discoveries}.

In summary, we make the following contributions:
\begin{itemize}
  \item We show that both sequential and parallel autoresearch agent scaling paradigms plateau under a fixed budget (Section~\ref{sec:exp-plateau}). We further show that this plateau happens because the agent becomes more conservative as its session grows longer (Section~\ref{sec:analysis}).
  \item We propose \emph{forgetting}: periodically discarding every agent's session and the candidate history and restarting from the best candidate found so far (Section~\ref{sec:approach}).
  \item We combine forgetting with a team of agents with communication, evaluate it on four diverse tasks, and find steady improvement over strong parallel-agent baselines, by 10\% on average (Section~\ref{sec:experiments}).
\end{itemize}

\section{Related Work}\label{sec:related}

\paragraph{Autoresearch.}

Hillclimbing with language models began with a single model call as
the proposer inside an evolutionary loop. FunSearch~\citep{romeraparedes2024funsearch} and AlphaEvolve~\citep{novikov2025alphaevolve} pair an LLM with an automated evaluator and a database of programs, and prompt optimizers~\citep{yang2024opro,agrawal2026gepa} follow the same pattern, with adaptive variants that route budget across populations as the search stalls~\citep{cemri2026adaevolveadaptivellmdriven} or evolve the search strategy itself~\citep{liu2026evoxmetaevolutionautomateddiscovery}. Each proposal in these systems is a fresh prompt, and diversity is maintained at the level of the population. Agents can also branch together with their workspace, where forking outperforms single paths at matched cost~\citep{tan2026parallelenv}. A more recent line of work hands the whole loop to a coding agent. ScientistOne~\citep{meng2026scientistone} runs the full cycle from hypothesis to experiment to write-up, Meta-Harness~\citep{lee2026metaharness} has an agent rewrite the harness around a model and keep the versions that score better, and autoresearch~\citep{karpathy2026autoresearch} reduces this to one agent that edits a training script in a single session and is told never to stop. However, recent studies also find that agents that inherit the history of failed attempts solve far fewer tasks than those that start clean~\citep{tan2025recoverybench} and such iterative loops lose diversity and stall~\citep{he2026beneathdiffdiagnosingmitigating, virk2026swarmresearchorchestratingcodingagents}.

Recent work also explores alternative ways to manage optimization state and the proposer agent. Stateless Language Agents~\citep{zhang2026statelesslanguageagentsscaling} use a stateless advisor that assigns each experiment to isolated workers, which coordinate through the harness state.
A concurrent work, Fork-and-Flush~\citep{cai2026forkandflushescapingideabasins}, periodically forks an agent into fresh-context trajectories that inherit the whole workspace. While the idea of resetting agent context sessions is similar, we more aggressively remove the workspace state as well, keeping only the best candidate solution. Additionally, our agents communicate with each other within each round before and after forgetting without an orchestrator.

\paragraph{Multi-Agent Systems.}
Multi-agent LLM systems organize collaboration through assigned roles and structured communication~\citep{wu2023autogenenablingnextgenllm, hong2024metagptmetaprogrammingmultiagent, qian-etal-2024-chatdev}, or drop the orchestrator and let agents coordinate through a shared log of findings and a task queue~\citep{mao2026decentralizedmultiagentsystemsshared}. Allowing agents to read or debate one another's outputs can improve accuracy on individual queries~\citep{du2024debate,wang2024mixtureofagentsenhanceslargelanguage,qian2025scalinglargelanguagemodelbased}, and teams with communication can outperform independent ones on open-ended optimization given clear feedback and enough compute~\citep{park2026scalingdiscoverytesttimecommunication}. Collaboration can still fail in systematic ways~\citep{cemri2025multiagentllmsystemsfail}, and the gain is conditional because communication has two effects on different timescales. It propagates improvements, which compounds over a long run with a verifier, and it homogenizes proposals, which happens the moment agents read one another's full solutions~\citep{ann2026interactiontaxcommunicationerases}. Under short budgets and matched compute the second effect dominates: teams often fail to beat a strong single agent~\citep{kim2026sciencescalingagentsystems}, homogeneous agents saturate as their outputs correlate~\citep{yang2026understandingagentscalingllmbased}, deliberation drifts to a compromise that holds back the best member~\citep{pappu2026multiagentteamsholdexperts}, and debating agents adopt an incorrect majority view~\citep{wynn2025talkisntcheapunderstanding}.

\section[Agent Teams That Forget]{Agent Teams That Forget}\label{sec:approach}

\begin{figure}[ht]
  \centering
  \includegraphics[width=\textwidth]{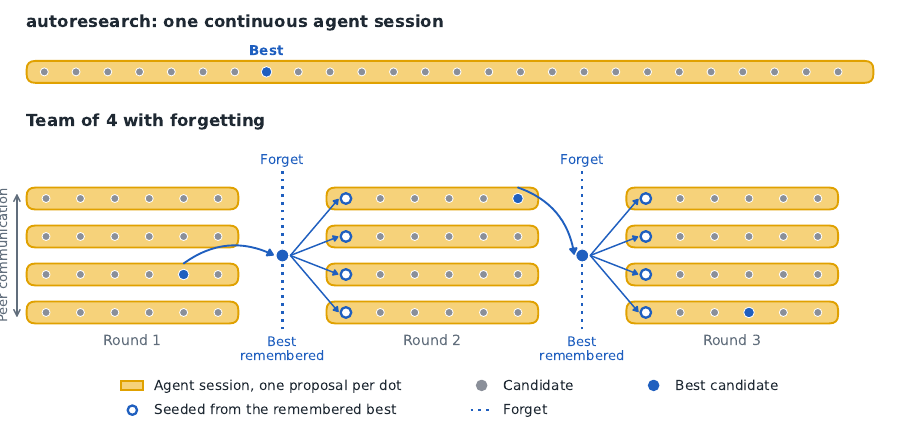}
  \caption{\textbf{The core idea.} Top, autoresearch, where one agent proposes candidates iteratively inside a single continuous session with the whole history in context. Running several such agents in parallel or letting them read one another's candidates through peer communication still keeps every session, and all of these configurations plateau. Bottom, a team of four agents with communication and forgetting. At each forget line every session and search state are discarded, and only the best candidate so far is remembered, so all four agents of the next round start fresh from that single candidate.}
  \label{fig:idea}
\end{figure}

Given a problem, an autoresearch agent improves the solution iteratively: the agent proposes a candidate, a judge evaluates it, and then the agent uses scores and feedback from the judge to decide what to try next. As this loop runs, two types of history accumulate. The \emph{search state} records every candidate together with its score and feedback, while the agent's \emph{session} holds the messages that produced these candidates, including the reasoning behind every attempt. Standard autoresearch, shown at the top of Figure~\ref{fig:idea}, keep both kinds of history within one continuous session. A team with forgetting, shown at the bottom, handles this history in a specific way by periodically discarding every session together with the search state and keeping only the best candidate found so far.

\subsection[Forgetting]{Forgetting}\label{sec:approach-forgetting}

Forgetting divides the search into rounds. At the end of each round, the agent discards its session together with the search state and keeps only the best candidate found so far along with its score. A new agent then starts the next round from this candidate with an empty context and the same instructions as the first round. We call each such step a \emph{forget}. In Figure~\ref{fig:idea}, each forget line marks the end of a round.

We keep only the best candidate because it already contains most of the progress of the previous agent sessions, while the discarded states mainly record how that progress was made. The new agent is no longer conditioned on earlier failed attempts towards certain optimization directions. Our observation shows that with a fresh context and search state, the autoresearch agent produces more thinking tokens and much larger edits (Section~\ref{sec:analysis}).

\subsection[Teams]{Teams}\label{sec:approach-teams}

We implement a simple communication scheme for our agent teams. A team consists of several agents that work on the same problem in parallel, each in its own session and with its own search state. Before each proposal, every agent gets access to other agents' candidates as commits along with their scores. There is no direct, point-to-point communication between agents during the agent sessions. Therefore, communication is fully asynchronous and token-free. While this scheme is simple, we find that it works particularly well in our setting.

Forgetting combines naturally with agent teams. Communication lets a good idea found by one agent reach the others within a few proposals. Yet each agent still works within its own growing session, so every agent stalls in the same way as a single agent. Similarly, we extend forgetting to the whole team. At each forget, every agent discards its session and search state, and all agents restart from the best candidate across the team. The bottom of Figure~\ref{fig:idea} shows this for a team of four. We discuss the benefit of combining forgetting with agent teams in Section~\ref{sec:exp-ablations}.

\subsection[Implementation]{Implementation}\label{sec:approach-implementation}

\begin{figure}[t]
  \centering
  \includegraphics[width=\textwidth]{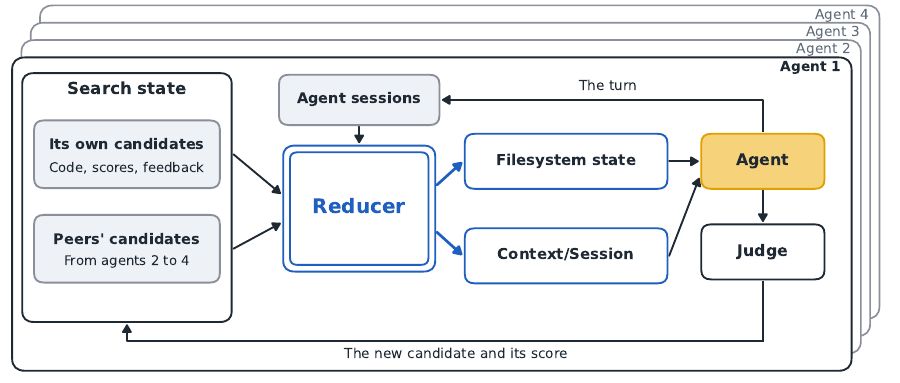}

  \caption{\textbf{The harness.} The reducer reads the agent's search state and its sessions and decides what the agent sees and which session it continues. The agent then edits a candidate and the judge scores it. The harness records the new candidate with its score in the search state and the turn in the agent sessions. A team runs four such loops at once, each with its own state, and the candidates of an agent's peers come from the other three loops.}
  \label{fig:harness}
\end{figure}

We implement agent teams with \emph{forgetting} in a small harness, where each agent runs the loop of Figure~\ref{fig:harness}. The search state of an agent is an append-only record of its candidates, where every candidate is a commit in a version-controlled repository together with the judge's score and feedback. The harness keeps the session script of every past agent session.

The central component is the \emph{reducer}, which controls the search state and the agent sessions of the agent's next turn. It decides which candidates the agent sees as files in its workspace and which session the agent continues. The agent then edits a candidate using only file tools, and the harness records the result and runs the judge on it. Every configuration we compare is a setting of the reducer: a single agent and the parallel agents see only their own candidates and continue their latest session, whereas a team also sees the candidates of its peers. Forgetting clears both kinds of memory. The harness starts every agent on a new search state that holds only the best candidate, so its workspace shows no earlier candidates. With no past session to continue, the agent also starts with an empty context.

\section[Experiments]{Experiments}\label{sec:experiments}

We evaluate on four tasks. AtCoder Heuristic Contest 014 (AHC014) and Frontier-CS Q171 are combinatorial optimization problems. The Anthropic kernel performance challenge optimizes a kernel for a simulated VLIW SIMD machine. Finally, nanochat-autoresearch improves the training of a small language model in five minutes. Appendix~\ref{sec:app-tasks} describes each task and its budget in detail (Table~\ref{tab:tasks}). We run all our experiments with the Pi agent \citep{zechner2025pi} and the DeepSeek V4.1 Flash model \citep{deepseekai2026deepseekv41flash}.

We evaluate the four tasks in three settings, a single agent, four parallel agents and a team with forgetting. All three settings use the same model and the same budget. The single agent continues one session for the whole budget, and the parallel agents split the budget without seeing one another's work. The team with forgetting has four agents that communicate through their candidates. It forgets three times, which splits its budget into four rounds. We ran each configuration three times per task.

\subsection[Parallel Scaling and Communication Still Plateau]{Parallel Scaling and Communication Still Plateau}\label{sec:exp-plateau}

\begin{table}[t]
  \centering
  \small
  \setlength{\tabcolsep}{4pt}\renewcommand{\arraystretch}{1.1}
\begin{tabular*}{\textwidth}{@{\extracolsep{\fill}}l r r@{\extracolsep{0pt}\,$\pm$\,}l@{\extracolsep{\fill}\hspace{\tabcolsep}} r r@{\extracolsep{0pt}\,$\pm$\,}l@{\extracolsep{\fill}\hspace{\tabcolsep}}}
  \toprule
  & \multicolumn{3}{c}{AHC014 score ($\times 10^3$) {\color{better}$\uparrow$}} & \multicolumn{3}{c}{VLIW kernel cycles {\color{better}$\downarrow$}} \\
  \cmidrule(lr){2-4} \cmidrule(lr){5-7}
  Method & Best\makebox[2.4em]{} & \multicolumn{2}{c}{Mean $\pm$ s.d.\makebox[2.4em]{}} & Best\makebox[2.4em]{} & \multicolumn{2}{c}{Mean $\pm$ s.d.\makebox[2.4em]{}} \\
  \midrule
  \textcolor{single}{1 agent} & 961 \makebox[2.4em][l]{} & 795 & 210 \makebox[2.4em][l]{} & 1,262 \makebox[2.4em][l]{} & 1,426 & 197 \makebox[2.4em][l]{} \\
  \textcolor{parallel}{4 parallel agents} & 1,007 \makebox[2.4em][l]{{\color{better}\scriptsize(+5\%)}} & 964 & 57 \makebox[2.4em][l]{{\color{better}\scriptsize(+21\%)}} & 1,172 \makebox[2.4em][l]{{\color{better}\scriptsize(-7\%)}} & 1,220 & 44 \makebox[2.4em][l]{{\color{better}\scriptsize(-14\%)}} \\
  \textcolor{ours}{\textbf{Team with forgetting}} & \textbf{1,145} \makebox[2.4em][l]{{\color{better}\scriptsize(+19\%)}} & \textbf{1,103} & 50 \makebox[2.4em][l]{{\color{better}\scriptsize(+39\%)}} & \textbf{1,077} \makebox[2.4em][l]{{\color{better}\scriptsize(-15\%)}} & \textbf{1,097} & 18 \makebox[2.4em][l]{{\color{better}\scriptsize(-23\%)}} \\
  \bottomrule
\end{tabular*}

  \vspace{1.5ex}
  \setlength{\tabcolsep}{4pt}\renewcommand{\arraystretch}{1.1}
\begin{tabular*}{\textwidth}{@{\extracolsep{\fill}}l r r@{\extracolsep{0pt}\,$\pm$\,}l@{\extracolsep{\fill}\hspace{\tabcolsep}} r r@{\extracolsep{0pt}\,$\pm$\,}l@{\extracolsep{\fill}\hspace{\tabcolsep}}}
  \toprule
  & \multicolumn{3}{c}{Frontier-CS Q171 score {\color{better}$\uparrow$}} & \multicolumn{3}{c}{nanochat-autoresearch val BPB {\color{better}$\downarrow$}} \\
  \cmidrule(lr){2-4} \cmidrule(lr){5-7}
  Method & Best\makebox[2.4em]{} & \multicolumn{2}{c}{Mean $\pm$ s.d.\makebox[2.4em]{}} & Best\makebox[2.4em]{} & \multicolumn{2}{c}{Mean $\pm$ s.d.\makebox[2.4em]{}} \\
  \midrule
  \textcolor{single}{1 agent} & 68.2 \makebox[2.4em][l]{} & 64.0 & 6.2 \makebox[2.4em][l]{} & 0.977 \makebox[2.4em][l]{} & 0.980 & 0.003 \makebox[2.4em][l]{} \\
  \textcolor{parallel}{4 parallel agents} & 71.4 \makebox[2.4em][l]{{\color{better}\scriptsize(+5\%)}} & 70.2 & 1.1 \makebox[2.4em][l]{{\color{better}\scriptsize(+10\%)}} & 0.979 \makebox[2.4em][l]{{\color{worse}\scriptsize(+0.2\%)}} & 0.980 & 0.001 \makebox[2.4em][l]{{\color{better}\scriptsize(-0.1\%)}} \\
  \textcolor{ours}{\textbf{Team with forgetting}} & \textbf{82.8} \makebox[2.4em][l]{{\color{better}\scriptsize(+21\%)}} & \textbf{80.8} & 2.9 \makebox[2.4em][l]{{\color{better}\scriptsize(+26\%)}} & \textbf{0.954} \makebox[2.4em][l]{{\color{better}\scriptsize(-2\%)}} & \textbf{0.961} & 0.012 \makebox[2.4em][l]{{\color{better}\scriptsize(-2\%)}} \\
  \bottomrule
\end{tabular*}

  \caption{Final best score of each configuration, over three runs per task, with the same total token cost. Bold marks the best configuration on each task; percentages are the change relative to the single agent baseline.}
  \label{tab:results}
\end{table}

\begin{table}[t]
  \centering
  \small
  \setlength{\tabcolsep}{4pt}\renewcommand{\arraystretch}{1.1}
\begin{tabular*}{\textwidth}{@{\extracolsep{\fill}}l r@{\hspace{0.4em}}l r@{\hspace{0.4em}}l r@{\hspace{0.4em}}l r@{\hspace{0.4em}}l}
  \toprule
  & \multicolumn{2}{c}{AHC014 ($\times 10^3$) {\color{better}$\uparrow$}} & \multicolumn{2}{c}{VLIW cycles {\color{better}$\downarrow$}} & \multicolumn{2}{c}{Frontier-CS Q171 {\color{better}$\uparrow$}} & \multicolumn{2}{c}{nanochat-autoresearch {\color{better}$\downarrow$}} \\
  \midrule
  Before forget & 931 &  & 1,231 &  & 73.2 &  & 0.975 &  \\
  After forget 1 & 1,035 & {\color{better}\scriptsize(+11.2\%)} & 1,145 & {\color{better}\scriptsize(-7.0\%)} & 75.8 & {\color{better}\scriptsize(+3.6\%)} & 0.965 & {\color{better}\scriptsize(-1.0\%)} \\
  After forget 2 & 1,085 & {\color{better}\scriptsize(+4.8\%)} & 1,115 & {\color{better}\scriptsize(-2.6\%)} & 79.4 & {\color{better}\scriptsize(+4.7\%)} & 0.964 & {\color{better}\scriptsize(-0.1\%)} \\
  After forget 3 & 1,103 & {\color{better}\scriptsize(+1.7\%)} & 1,097 & {\color{better}\scriptsize(-1.6\%)} & 80.8 & {\color{better}\scriptsize(+1.8\%)} & 0.961 & {\color{better}\scriptsize(-0.3\%)} \\
  \bottomrule
\end{tabular*}

  \caption{Mean best score of the team with forgetting at the end of each round, over its three runs per task, with the change from the previous round. Most of the gain comes from the first forget. As the candidate solution approaches the optimum, progress slows down.}
  \label{tab:rounds}
\end{table}

Running four agents in parallel helps on three of the four tasks (Table~\ref{tab:results}). For example, on AHC014 the parallel agents raise the mean best score by 21\% over a single agent. On nanochat-autoresearch they bring no gain at all. Parallel agents also slow down as their sessions grow (Figure~\ref{fig:main}). For example, over the second half of the budget their mean best score on AHC014 improves by only 1.2\%, and their cycle count on the VLIW kernel falls by only 2.4\%. Frontier-CS Q171 is the exception, where parallel agents keep improving, but the team with forgetting still ends 15\% above them.

Communication alone does not remove the plateau either. Before its first forget, the team is simply a group of agents with communication that keep their sessions. On some tasks communication already helps at this point. For example, on Frontier-CS Q171 the team leads the parallel agents by 13\% at the same spend, while on AHC014 it is slightly behind them. Still, communication does not prevent the plateau. As we observed in Figure~\ref{fig:main}, each team configuration's progress still flattened before each forget.
In an additional experiment in Section~\ref{sec:exp-ablations}, continuing the sessions of a team with communication for 30 more proposals per agent gains only 1.4\%.

\subsection[Forgetting Unblocks the Plateau]{Forgetting Unblocks the Plateau}\label{sec:exp-results}

The team with forgetting achieves the best result on every task, both on average and in its best run (Table~\ref{tab:results}). On average across the four tasks, it improves on the parallel agents by 10\% and on a single agent by 22\%. For example, on AHC014 its mean best score is 14\% above the parallel agents and 39\% above a single agent. On the VLIW kernel it also has the smallest spread of the three configurations, being 10\% better than the parallel agents and 23\% better than a single agent. 

Table~\ref{tab:rounds} shows these gains over the rounds. The mean best score improves after every forget on every task, and 32 of the 36 forgets lead to a new best candidate within the next round. The gain from each forget generally shrinks over the rounds. For example, on AHC014 the mean best score rises by 11.2\% after the first forget and by 1.7\% after the third. This is expected, since most of our tasks have an explicit or implicit upper bound, and progress becomes harder as the search approaches it.

\subsection[Forgetting Outperforms Continuing at Lower Cost]{Forgetting Outperforms Continuing at Lower Cost}\label{sec:exp-ablations}

\begin{figure}[tb]
  \centering
  \includegraphics[width=\textwidth]{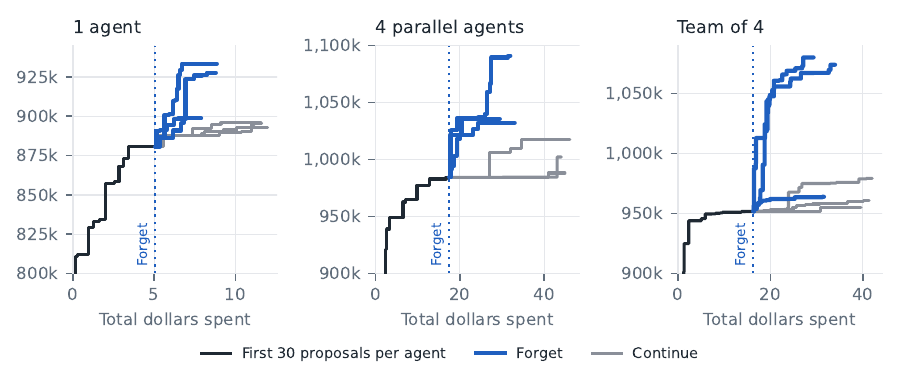}
  \caption{\textbf{Continue or forget, from the same plateaued state.} Best score so far on AHC014 against total money spent. The black line shows the first 30 proposals prefix for all runs. At the forget line, the next 30 proposals per agent either continue the same sessions or forget. Left to right, a single agent, four parallel agents and a team of four. Forgetting climbs higher on average and spends less than continuing the same sessions overall due to shorter context.}
  \label{fig:ablation}
\end{figure}

To understand whether forgetting helps across agent setups, we start from a run that has already plateaued and spend its next round in two ways. \emph{Continue} keeps every session and candidate, as if the plateaued agent continues in place. \emph{Forget} discards all the states and sessions and keeps only the best candidate. Both then run 30 more proposals per agent. We do this on AHC014 for a single agent, four parallel agents and a team of four (Figure~\ref{fig:ablation} and Table~\ref{tab:ablation}).

Continuing barely moves any configuration. The single agent gains only 1.6\%, the parallel agents 1.9\% and the team 1.4\%, even though every continued turn costs more because it carries the long session. Forgetting gains 4.5\% for the single agent, 7.0\% for the parallel agents and 9.2\% for the team at little more than half the cost. 

Forgetting also works best with a team. After forgetting, the team gains 9.2\%, the highest across all three setups. Four fresh agents explore more directions from the best candidate than one, and communication broadcasts a good direction found by one of them to the others within the same round. We show one concrete example of why this works in Section~\ref{sec:analysis}.

\begin{table}[tb]
  \centering
  \small
  \setlength{\tabcolsep}{8pt}\renewcommand{\arraystretch}{1.1}
  \begin{tabular}{l l r r}
    \toprule
    Starting state & Next 30 proposals & Gain over the start & Spend \\
    \midrule
    \multirow{2}{*}{\textcolor{single}{1 agent}, best 880,753} & Continue & +1.6\% $\pm$ 0.2\% & \$6.66 \\
    & \textcolor{ours}{\textbf{Forget}} & \textbf{+4.5\% $\pm$ 2.1\%} & \$3.44 \\
    \addlinespace
    \multirow{2}{*}{\textcolor{parallel}{4 parallel agents}, best 984,380} & Continue & +1.9\% $\pm$ 1.5\% & \$27.51 \\
    & \textcolor{ours}{\textbf{Forget}} & \textbf{+7.0\% $\pm$ 3.3\%} & \$13.97 \\
    \addlinespace
    \multirow{2}{*}{\textcolor{ours}{Team of 4}, best 951,931} & Continue & +1.4\% $\pm$ 1.3\% & \$24.67 \\
    & \textcolor{ours}{\textbf{Forget}} & \textbf{+9.2\% $\pm$ 6.9\%} & \$15.32 \\
    \bottomrule
  \end{tabular}
  \caption{Continuing versus forgetting from the same plateaued state on AHC014. Gain is the best new candidate over the state's best, as mean and standard deviation over three replicates. Spend is per replicate.}
  \label{tab:ablation}
\end{table}

\section[Why forgetting helps]{Why Forgetting Helps}\label{sec:analysis}

\begin{figure}[t]
  \centering
  \includegraphics[width=\textwidth]{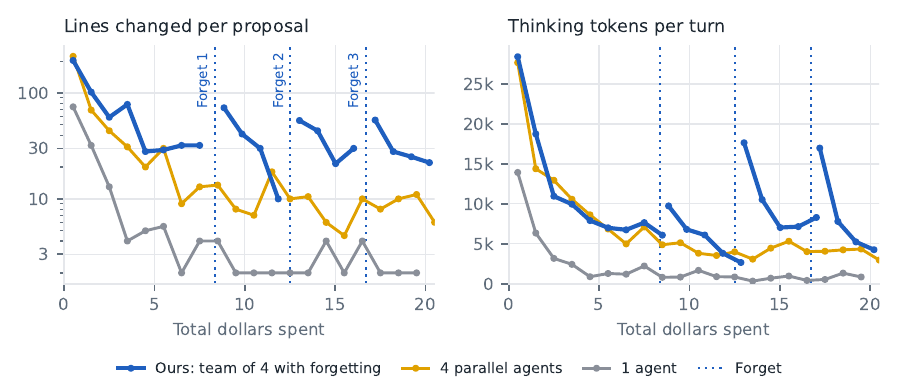}
  \caption{\textbf{Every session narrows, and forgetting resets it.} Inside the AtCoder Heuristic Contest 014 runs of Table~\ref{tab:results}, three replicates per configuration, pooled. Each point is the median over the proposals that finished within one dollar of a replicate's spend; the team's line breaks at each forget, where its sessions are discarded and it restarts from the best candidate.}
  \label{fig:analysis}
\end{figure}

We analyze the agent trajectories behind Figure~\ref{fig:main} and Table~\ref{tab:results} to understand how search changes as sessions grow and what changes after forgetting. Figure~\ref{fig:analysis} tracks code changes and reasoning tokens per proposal. 

\paragraph{Search Narrows as Sessions Grow}
As sessions lengthen, agents make smaller edits and use less reasoning. In Figure~\ref{fig:analysis}, all three configurations begin with edits of around 100 lines and then decrease within the first few dollars. Reasoning tokens also falls to a third or less of its initial level. By the fourth dollar, the single agent only edits a few lines. Before its first reset, the team's edits fall from about 150 lines to 30 and reasoning from about 28k tokens to 7k, showing the same decline as its performance plateaus in Figure~\ref{fig:main}.

On VLIW, within the \$40 budget of each of the three single-agent replicates, 532 of 763 completed turns produce no candidate, consuming 66.2\% of the model spend. This behavior appears in all three replicates, consuming between 58.4\% and 70.4\% of their spend. 
These empty turns are not failures. In 531 of the 532, the agent reads at most one file, makes no edit and ends its turn by declaring a plateau. For example, after 34 turns at 1,710 cycles, one single agent writes \emph{``I've reached the practical limit of this machine configuration''} and proposes nothing. Of the agent's remaining 201 turns, 164 end the same way. Yet 1,710 cycles is far from the practical limit. The same agent later reaches 1,644 cycles, and the team with forgetting reaches 1,077 cycles on the same task.

The search can stall even when the agent continues to generate and consider new ideas. In one single-agent VLIW session, eleven consecutive turns produce no candidate even though the agent develops an untested selection-tree implementation. It initially rejects the idea using a mistaken comparison of its estimated cost with the kernel's load bound. A later turn explicitly corrects that comparison and works through the register requirements, but still declines to implement the change because it appears too complex and risky. The session thus stalls where the agent continues to generate ideas but no longer turns them into experiments.

\paragraph{Forgetting Renews Collective Search}
After forgetting, the agents resume larger edits and use more reasoning tokens per proposal. In AHC014, edits return to about fifty changed lines and reasoning-token usage rises to twice its pre-reset level or more. 
The runs behind the same-state experiment in Figure~\ref{fig:ablation} show a similar trend. Over the first ten proposals per agent, median edits are 40 lines after forgetting versus 6 under continuation, and median reasoning-token usage is 9.4k versus 2.9k. Both increases appear in all three fresh-team replicates, alongside the performance advantage in Table~\ref{tab:ablation}. Edit size and reasoning-token usage decline again within a fresh phase.

Forgetting and communication work well together. Before the first forget in one VLIW team run, the team's last improvements are only a few cycles each, taking its best from 1,212 to 1,201 cycles. After the forget, one fresh agent tries a much larger change, a new way to read values from the tree in the kernel. Its first attempt is slower at 1,409 cycles, and its second only matches the starting point. The agent then reads a peer's diff and borrows a small trick that simplifies its new code. The combined version reaches 1,136 cycles, the largest single improvement of the round, and the team ends the round at 1,114 cycles.  Without forgetting, the agent would stay conditioned on its accumulated context, keep making small changes and never try the bold idea. Without communication, the agent would give up on the idea after its two failed attempts, without absorbing a fix from its peer.

\section[Conclusion]{Conclusion}\label{sec:conclusion}

In conclusion, we propose \emph{forgetting}, a general mechanism for long-running discovery agents. Forgetting addresses the plateau problem by discarding the sessions and the search state and keeping only the best candidate periodically. 

Combined with a team of agents with communication, forgetting gives the best result on all four of our tasks and improves on parallel agents by up to 15\% at the same budget. These results suggest forgetting pushes the performance of autoresearch agents. A natural next step is to decide when to forget and what to remember.

\section*{Acknowledgement}

This research is supported by gifts from Accenture, Algorithmic SuperIntelligence Labs, Amazon, AMD, Anyscale, Broadcom, cmpnd, Google, IBM, Intel, Intesa Sanpaolo, Lambda, Lightspeed, Mirendil, NVIDIA, Soma Labs, and VESSL.

\bibliography{references}
\bibliographystyle{colm2026_conference}

\newpage
\appendix
\section[Task Descriptions]{Task Descriptions}\label{sec:app-tasks}\label{sec:appendix}

\begin{table}[H]
  \centering
  \small
  \begin{tabularx}{\textwidth}{@{}l >{\raggedright\arraybackslash}p{2.3cm} >{\raggedright\arraybackslash}X r@{}}
    \toprule
    Task & Candidate & Problem and objective & Budget \\
    \midrule
    AHC014 & C++ program & On a grid with initial dots, repeatedly add a dot that completes a rectangle with three existing dots. Rectangles are axis-aligned or tilted by 45 degrees, and their sides may not pass through other dots or overlap earlier rectangles. Maximize the mean score over 30 official test cases, where dots far from the center count more. & \$20 \\
    \addlinespace
    VLIW kernel & Python program & Write a kernel for a simulated VLIW SIMD machine that traverses a tree and hashes a batch of values, emitted as bundles of parallel instructions. Minimize the clock cycles of the kernel, which start at 147,734 for the reference implementation. & \$40 \\
    \addlinespace
    \begin{tabular}[t]{@{}l@{}}Frontier-CS\\Q171\end{tabular} & C++ program & On a $20 \times 20$ grid, visit 40 target squares in a given order by moving one square, sliding until blocked, and placing or removing blocks. Use as few actions as possible. Maximize the mean score over 75 test cases, scaled per case from 0 for a greedy baseline to 100 for a reference solution. & \$20 \\
    \addlinespace
    \begin{tabular}[t]{@{}l@{}}nanochat-\\autoresearch\end{tabular} & PyTorch training script & Train a small language model for 5 minutes on one H100. Minimize the validation bits per byte (val BPB). & \$20 \\
    \bottomrule
  \end{tabularx}
  \caption{Tasks. Each task gives the agent a candidate to edit and a judge that scores it; the budget is the money spent per replicate of every treatment.}
  \label{tab:tasks}
\end{table}

\section[Prompts]{Prompts}\label{sec:app-prompts}

We deliberately keep the prompts simple. The prompts describe the task and the loop, and they give the agent no strategy for the search itself. Every agent is a Pi agent with its default system prompt. The harness then sends one message per turn. The first turn of a session receives the task, the current status, and the instructions below. Every later turn of the same session receives only the current status and a short reminder. After a forget, the new session again receives the full first-turn message and starts from the remembered candidate. The candidates of the agent and its peers are files in the agent's workspace, which it reads with its own tools. The lines about peers appear only for a team, one line per peer.

\paragraph{First turn of a session.}
\begin{Verbatim}[frame=single,fontsize=\footnotesize,framesep=3mm]
# Task

<the task description and its rules (Appendix A)>

# Feedbacks and Updates

Round <k>.
Your current candidate is `<id>`, mean score <score>.
Branch <i>: best `<id>` at <score>; its pool is under `.context/peers/branch<i>/`.

# Instructions

`.context/` holds what was prepared for this turn: the candidate's feedback, its
lineage, other candidates, or the whole run. Read it before changing anything.
Every candidate is a commit in this repository: `git show <id>:<file>` shows any
of them, `git diff <a> <b>` compares two.

Improve the candidate, then end the session and say what you changed. Do not
evaluate the change yourself: another process scores your candidate after you
stop and invokes you again with the result, so a simulator or judge built here
only spends this session. Do not commit; the harness commits your working tree,
scores it, and continues from it in a later turn. Work inside this directory
only: reads and writes outside it are refused, and there is no need for
background tasks. Anything you leave here (tests, scripts) is committed with your
change and is there next turn.

Once your candidate is in place, end the session. Another process evaluates it
and invokes you again.
\end{Verbatim}

\paragraph{Every later turn of the same session.}
\begin{Verbatim}[frame=single,fontsize=\footnotesize,framesep=3mm]
# Feedbacks and Updates

Round <k>.
Your current candidate is `<id>`, mean score <score>.
Branch <i>: best `<id>` at <score>; its pool is under `.context/peers/branch<i>/`.

# Continue

The judge's feedback and the run's state are in `.context/` as before; TASK.md
holds the full instructions. Improve the candidate, then end the session. Another
process scores it and invokes you again.
\end{Verbatim}

\end{document}